\documentclass{article}

\usepackage{arxiv}

\usepackage[utf8]{inputenc} 
\usepackage[T1]{fontenc}    
\usepackage{hyperref}       
\usepackage{url}            
\usepackage{booktabs}       
\usepackage{amsfonts}       
\usepackage{nicefrac}       
\usepackage{microtype}      
\usepackage{lipsum}
\usepackage{graphicx}
\usepackage{graphicx}
\usepackage{siunitx}
\usepackage{tikz}
\usepackage{mathtools}
\usepackage{bm}
\usepackage{natbib}
\graphicspath{ {./images/} }

\newcommand{\greencheck}{}%
\DeclareRobustCommand{\greencheck}{%
  \tikz\fill[scale=0.4, color=green]
  (0,.35) -- (.25,0) -- (1,.7) -- (.25,.15) -- cycle;%
}

\title{GRASP EARTH: Intuitive Software for Discovering Changes on the Planet}

\author{
 Waku Hatakeyama \\
  Ridge-i Inc.\\
  \texttt{whatakeyama@ridge-i.com} \\
   \And
 Shirou Kawakita \\
  Ridge-i Inc.\\
  \texttt{skawakita@ridge-i.com} \\
  \And
 Ryohei Izawa\\
  Ridge-i Inc. \\
  \texttt{rizawa@ridge-i.com} \\
  \And
  Masanari Kimura \\
  Ridge-i Inc. \\
  \texttt{mkimura@ridge-i.com}
}

\begin{document}

\maketitle 

\begin{abstract}
Detecting changes on the Earth, such as urban development, deforestation, or natural disaster, is one of the research fields that is attracting a great deal of attention.
One promising tool to solve these problems is satellite imagery.
However, satellite images require huge amount of storage, therefore users are required to set Area of Interests first, which was not suitable for detecting potential areas for disaster or development.
To tackle with this problem, we develop the novel tool, namely GRASP EARTH, which is the simple change detection application based on Google Earth Engine.
GRASP EARTH allows us to handle satellite imagery easily and it has used for disaster monitoring and urban development monitoring.
\end{abstract}


\section{Introduction}
GRASP EARTH is a series of applications which are developed on Google Earth Engine (GEE)~\citep{gorelick2017google}.
GEE was launched on 2010 and it has been served as a cloud computing platform for satellite images. 
Recently, a number of applications ranging from deforestation monitoring to disaster monitoring are created on this platform~\citep{amani2020google}.
Data availability and integrated code developing tool are the main reasons why a number of applications have been created. 
Among all the applications, Landsat and Sentinel are the most used data~\citep{amani2020google}.
Sentinel-1~\citep{torres2012gmes} and Sentinel-2~\citep{drusch2012sentinel} can acquire Synthetic Aperture Radar (SAR) images and optical images for each and they show relatively high resolution among the freely available images. 
Due to their plenty archived data, now users are ready for conducting not only single-time analysis but also time-series analysis.


By utilizing GEE, we developed a family of web application, namely "GRASP EARTH".
GRASP EARTH consists of several applications which combine comprehensive change detection engine and effective visualizing interface.


\subsection{Optical and SAR images}
Now, the number of satellite data which we can access to has been increasing. 
Among those sensors, optical data and Synthetic Aperture Radar (SAR) data have been frequently used because of its applicability. 
Optical satellites observe the reflected sun light from the surface of the Earth. 
Typical optical sensors can detect multiple spectral bands which range from about \SI{400}{\nano\meter} to \SI{1000}{\nano\meter}. 
Since the vegetation tends to have a unique spectral profile in infra-red band, optical data is widely used for precision agriculture and deforestation monitoring in the tropical rain forests. For example, \cite{hansen2013high} evaluated global forest change between 2000 and 2012 using 30m resolution optical images.

While optical satellites use the reflected light from Sun, SAR satellites emit microwave by themselves and observe the reflected signals after interacting with the surface of the Earth. Since the sensor uses a wavelength ranging from centimeter to meter, which can see through clouds, SAR satellites can observe regardless of the weather. 
Typically, the more number of artifacts exist in the place, the larger the intensity of returned signals become~\citep{matsuoka2004use}.
For example, the intensity of the returned signal of the urban area is larger than that of forest area.
Another example of SAR data is that large part of its signal reflect on the surface of the water, therefore the returned signals become less when the land is covered with water.
This feature is applied for the flood detection~\citep{shahabi2020flood}. 


\section{Overview of GRASP EARTH}\label{sec:overview}

GRASP EARTH consists of several applications which focus on detecting changes in different targets. 

GRASP EARTH Building\footnote{\url{https://ridgei.users.earthengine.app/view/grasp-earth-building-en}} 
is one of the applications which based on Sentinel-1 (SAR satellite). 
In this application, backscatter coefficient is statistically processed and significantly changed areas are highlighted. 
Therefore, users can easily detect the area where development is progressing. This information can be used for development plan in urban areas. In addition, time-series transition graph can be drawn and users can recognize when the change occurred.

Overview of the algorithm is as follows: Let $\bm{A}_\text{c, dateX}$ and $\bm{A}_\text{d, dateX}$ is sampled areas where new artifacts were constructed and destructed for each. And $\text{dateX}$ means assigned date1 or date2 on control panel.
\begin{align}
    \text{threshold}_\text{blue} &= F_\text{otsu}(\bm{A}_\text{c, date2} - \bm{A}_\text{c, date1}) \\
    \text{threshold}_\text{red} &= F_\text{otsu}(\bm{A}_\text{d, date2} - \bm{A}_\text{d, date1}) \\
    \text{Blue part} &= \bm{I}_\text{date2} - \bm{I}_\text{date1} \geq \text{threshold}_\text{blue} \\
    \text{Red part} &= \bm{I}_\text{date2} - \bm{I}_\text{date1} \leq \text{threshold}_\text{red} 
\end{align}
where, $\text{Blue part}$ and $\text{Red part}$ are blue area and red area in GRASP EARTH Building, $F_\text{otsu}$ is Otsu thresholding method~\citep{otsu1979threshold}.

GRASP EARTH Color\footnote{\url{https://ridgei.users.earthengine.app/view/grasp-earth-color-en}} 
is Sentine-2 (optical satellite) based application which enables users to compare two different dated RGB images at the same area. 
Optical satellites often suffer from the effect of clouds. 
To tackle with this problem, we developed a method to create cloudless images using several images taken in different dates.
First, we collect several images between two weeks before and two weeks after the assigned date. 
As a preprocessing, pixels which contain clouds are removed for each image. Lastly, median filter is applied along the time direction.

Overview of the algorithm is as follows:
\begin{align}
    \bm{I}_\text{out} = F_\text{median}(\bm{I}\in\mathbb{R}^{d} \otimes \bm{M}_\text{cloud}\in\mathbb{R}^{d})
\end{align}
where, $I_\text{out}$ is the output, $F_\text{median}$ is the median filter, $I$ is the input image, $M_\text{cloud}$ is the cloud mask and $\otimes$ is the operations of mask.

GRASP EARTH Pumice viewer and GRASP EARTH Pumice detector are the derived version of GRASP EARTH Color. In mid-August, Fukutokuokanoba undersea volcano erupted and generated a large amount of pumice. The pumice raft reached the coast of Okinawa, Japan on October 2021, which have plagued the fishing, tourism and industries there. 
Dedicated visualizer and detectors, namely GRASP EARTH Pumice viewer\footnote{\url{https://ridgei.users.earthengine.app/view/pumice-viewer-en}} 
and GRASP EARTH Pumice detector\footnote{\url{https://ridgei.users.earthengine.app/view/pumice-detector-en}} 
were developed. 
Similar to GRASP EARTH Color, Pumice viewer separates windows into two and shows two different RGB images taken by Sentinel-2. 
Default area focuses on where the pumice rafts are easily detected, but users can see other areas without defining AoI. 
GRASP EARTH Pumice detector was also developed.
In this application, as shown in eq.\ref{eq:ndwi}, first Normalized Difference Water Index (NDWI) is calculated to eliminate sea area. 
Since the brightness value and shapes of floating pumice might change, we adapted simple Otsu thresholding method and it is set as default value. 

Overview of the algorithm is as follows: Let $A_\text{p}$ is sampled pumice area, $A_\text{np}$ is not sampled pumice area.
\begin{align}
    \text{threshold} &= F_\text{otsu}(F_\text{ndwi}(A_\text{p})\cup F_\text{ndwi}(A_\text{np})) \label{eq:ndwi}\\
    \bm{I}_\text{pumice} &= (\bm{I}\in\mathbb{R}^{d} < \text{threshold})\otimes \bm{M}_\text{cloud}\in\mathbb{R}^{d}\otimes \bm{M}_\text{land}\in\mathbb{R}^{d}
\end{align}
where, $I_\text{out}$ is the output, $F_\text{median}$ is the median filter, $I$ is the input image, $M_\text{cloud}$ is the cloud mask, $M_\text{cloud}$ is the land mask and $\otimes$ is the operation of mask.

Default value was calculated by sampling about a hundred of pumice pixels and sea pixels of images over a few days on October 2021, and applied Otsu thresholding method after cloud masking so that clouds are not misidentified as pumice.
Simple but easily understandable user interface was a key factor of applying for actual business.
Detailed explanation is written in Appendix.

As shown in left picture of Figure~\ref{fig:conventional_vs_GE}, satellite data users have conventionally been required to set their area of interests (AoI) and specific time of observation before buying or searching satellite images.
There are several drawbacks in this flow. 
For instance, when we conduct a disaster monitoring, AoI is not usually set at the beginning.
Moreover, illegal deforestation and the other environmental crises also occur everywhere in the globe.
GRASP EARTH is designed to notice this kind of unpredictable events.

\begin{figure}[t]
    \centering
    \includegraphics[width=0.98\linewidth]{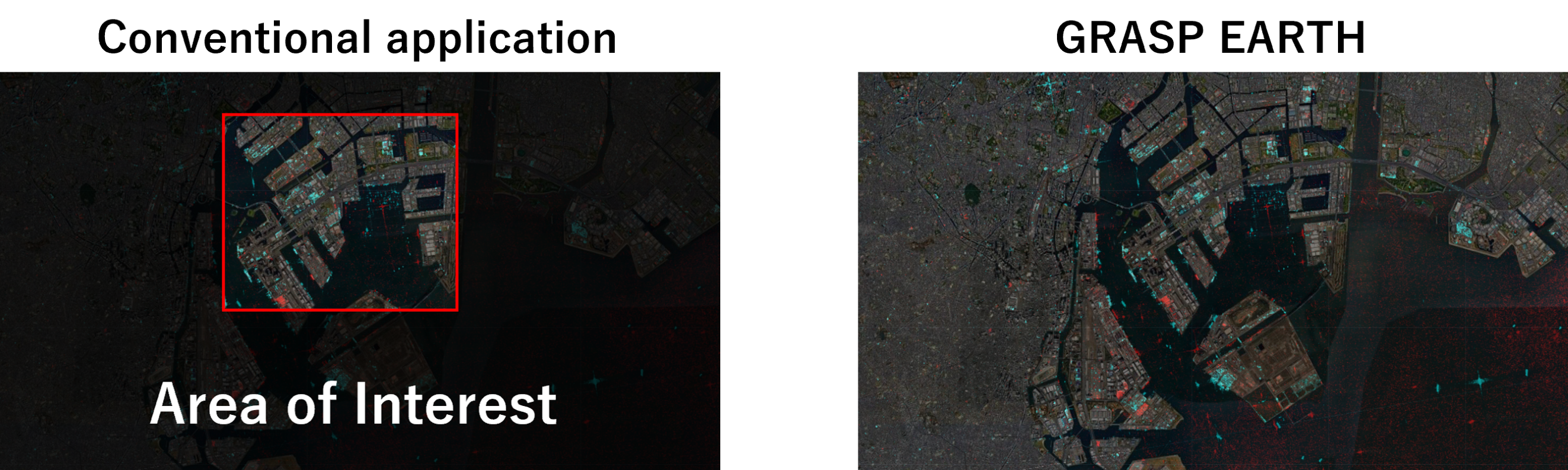}
    \caption{Comparison between the conventional applications and GRASP EARTH. Conventional applications needed to select the AoI, as shown in left picture, before analysis. On the other hand, GRASP EARTH helps users to find the AoI.}
    \label{fig:conventional_vs_GE}
\end{figure}

\subsection{How to use GRASP EARTH}

\begin{figure}[t]
    \centering
    \includegraphics[width=0.98\linewidth]{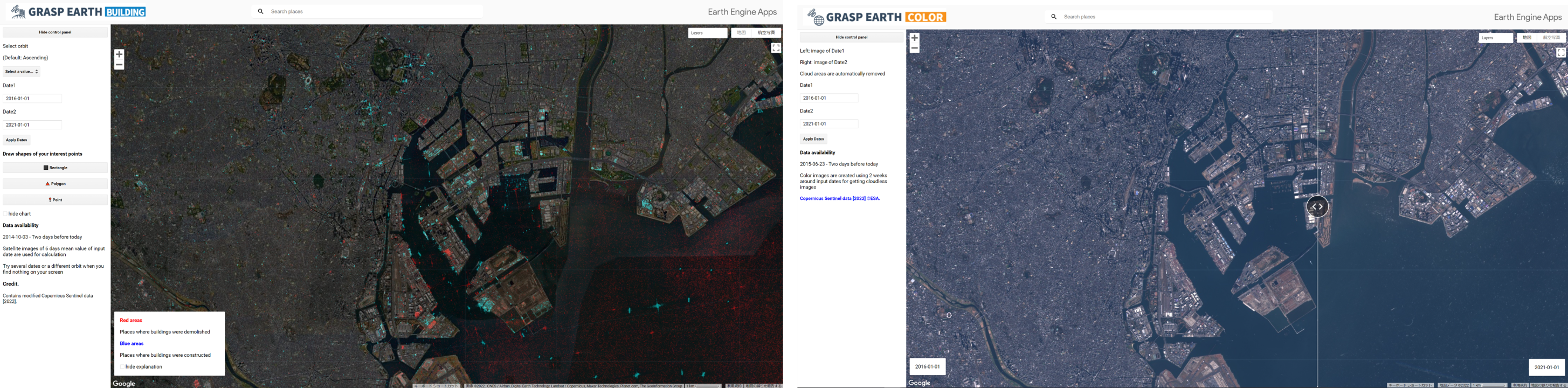}
    \caption{Left picture shows the screenshot of GRASP EARTH Building and right picture shows the screenshot of GRASP EARTH Color}
    \label{fig:GE_applications}
\end{figure}

Left picture of Figure~\ref{fig:GE_applications} shows the screenshot of GRASP EARTH Building looking at Tokyo bay, Japan. 
There are a number of blue areas and red areas drawn on the Google map aerial image.
Users can assign two different dates on the left control panel. 
Blue area shows the place where the value of backscatter coefficient increases, which means the artifacts are being built in urban area, while red area shows the place where the value of backscatter coefficient decreases, which means vice versa.
Users can also draw a point or a polygon on the map. After drawing, the graph of time-series average of backscatter coefficient over the drawn area comes out, which enables users to notice when the changes happened.

After identifying characteristics of changes around the world, users can check RGB images by using GRASP EARTH Color.
RGB images are created from Sentinel-2 observation. Right picture of Figure~\ref{fig:GE_applications} shows the screenshot of GRASP EARTH Color.
As GRASP EARTH Building, users can assign two different dates on the left control panel. 
Main window is split into two parts by movable splitter. 
The images of date1 and date2 is displayed on left side and right side of the window for each.



\section{Examples of GRASP EARTH usage}

\begin{figure}[t]
    \centering
    \includegraphics[width=0.90\linewidth]{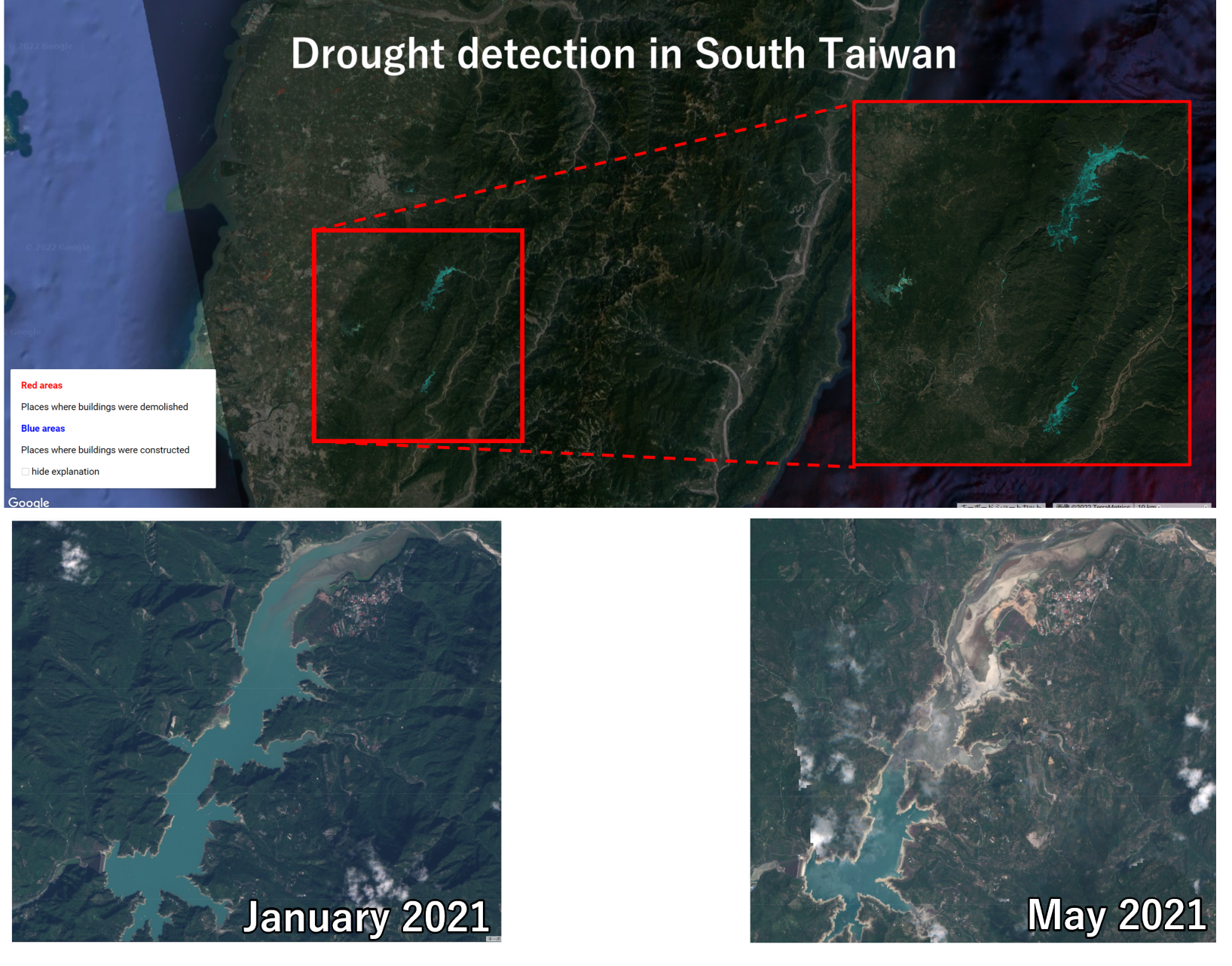}
    \caption{Top picture is the screenshot of GRASP EARTH Building which shows South Taiwan. Sentinel-1 images of January 2021 and May 2021 are compared. Several blue areas show where the dam dried up during the period. Bottom picture show the screenshot of GRASP EARTH Color which zoom in to Zengwen dam in South Taiwan on January 2021 and May 2021. North area of dam seems to be suffered from drought.}
    \label{fig:GE_taiwan}
\end{figure}

GRASP EARTH Building enabled users to estimate the effect of natural disaster. 
On May 2021, Taiwan experienced severe drought. 
During this period, several dams in Taiwan were suffered from low level of water storage. 
By using GRASP EARTH Building, users can find the dams which have low levels of water. 
Figure~\ref{fig:GE_taiwan} shows the screenshot of GRASP EARTH Building which shows South Taiwan. 
In this scenario, SAR images taken on January 2021 and May 2021 are processed. 
There are mainly three blue areas. 
Backscatter coefficient of water is lower than that of sand or mud, therefore, those dried areas are colored with blue.

Bottom part of Figure~\ref{fig:GE_taiwan} shows the screenshot of Zengwen dam located in South Taiwan. 
The north part of dam dried up on May 2021. 
Combining GRASP EARTH Building and GRASP EARTH Color, users can find and understand the environmental issue all over the world.

Conventional process of finding such an issue starts defining the AoI where drought occurred and analyze data. 
However, when we face with the natural disaster, it is difficult to define AoI first.
GRASP EARTH enables users to find the source of problems without defining AoI.

\section{Related works}
There are several existing change detection applications using satellite data which have user interfaces similar to our work ~\citep{kolodiy2020improvement, poortinga2018operational, scheip2021hazmapper, arevalo2020suite, kennedy2018implementation}. 
There are applications such as QGIS \footnote{\url{https://qgis.org/en/site/index.html}} and Monteverdi ~\footnote{\url{https://apollomapping.com/blog/free-for-all-monteverdi}} as viewers for satellite image data, and for change detection, there is Orfeo ToolBox ~\citep{grizonnet2017orfeo}, a satellite image processing library that provides algorithms that can be used in those applications. ERDAS IMAGINE \footnote{\url{https://www.hexagongeospatial.com/ja-jp/products/power-portfolio/erdas-imagine/erdas-imagine-remote-sensing-software-package}} also provides many types of change detection tools. Since those tools are native applications that run offline, it requires a large amount of optical and SAR image data to be downloaded beforehand in order to analyze a large area. 

Since the GEE was launched in 2010, it has been able to analyze a wide range of satellite image data at high speed. \cite{hansen2013high} mapped global deforestation and increase, and ~\cite{hird2017google} conducted a wetland analysis of a large area of Alberta. Another change detection applications with viewers are implemented on GEE ~\citep{poortinga2018operational, scheip2021hazmapper, arevalo2020suite, kennedy2018implementation} and those applications visualize changes in multiple bands such as NDVI, SWIR1, etc. 

There are also several change detection applications with viewers on other platforms such as Land Viewer ~\citep{kolodiy2020improvement} and Tellus-DEUCE, an add-in for Tellus-OS \footnote{\url{https://www.tellusxdp.com/en-us/os/}}. These applications extract changes after selecting the area to be investigated and visualize the changes.

\begin{table}[t]
\centering
\caption{Comparison between GRASP EARTH and existing applications. {\bf Unit} is "unit of time period", {\bf w/o AoI} is "without Area of Interests", {\bf Domain-agnostic} is whether the application focus on the specific domain.}
\label{undefined}
\begin{tabular}{cccccc}
               & {\bf Unit} & {\bf w/o AoI} & {\bf Optic} & {\bf SAR} & {\bf Domain-agnostic} \\ \hline
GRASP EARTH   & Date                & \greencheck           & \greencheck & \greencheck & \greencheck \\
LT-GEE \cite{kennedy2018implementation} & Date           & \greencheck           & \greencheck       & -- & \greencheck \\
visualize-ccdc \citep{arevalo2020suite} & Date                & \greencheck           & \greencheck       & -- & \greencheck \\
HazMapper \citep{scheip2021hazmapper}      & Date                & \greencheck           & \greencheck       & -- & -- \\
EcoDash \citep{poortinga2018operational} & Year                & \greencheck           & \greencheck       & -- & -- \\
Land Viewer \citep{kolodiy2020improvement}     & Date                & --                    & \greencheck       & -- & \greencheck \\
Tellus-DEUCE \footnote[8]   & Date                & --                    & \greencheck       & -- & \greencheck \\
\end{tabular}%
\end{table}

Although change detection applications on GEE do not require the selection of the area to be surveyed, there are issues such as the limited period of time over which changes can be visualized and the observation targets are limited to specific domains. Change detection applications on other platforms require the user to specify the area to be surveyed. GRASP EARTH is an application that visualizes changes in almost real time on a viewer that does not require the selection of pre-defined areas. Moreover, GRASP EARTH offers SAR based change detection algorithm. Therefore, it can be used even the surface of the Earth is covered with clouds.

\section{Conclusion and discussion}
Due to its low calculation cost, GRASP EARTH enables users to notice changes around the world with out setting AoI. Moreover, users can access to the latest data (up to two days before today). Though simple algorithms are implemented on the applications, they have used for tackling several problems.
If users want to detect pre-defined changes such as detecting the changes in the number of ships or cars, current version of GRASP EARTH should be improved. In this case, deep learning based change detection needs to be integrated. Since GEE platform offers Google Cloud Storage connection, integration with deep learning is easily achieved.

We can consider future works as follows. First, we can combine the method used in GRASP EARTH and the deep learning method. 
If this method is implemented, users can define AoI and monitoring duration quickly and proceed to dedicated analysis, which lead to reduce cost of purchasing satellite images and calculation time. 
Moreover, we can also consider to utilize text information on social network service, which can be also help detecting early symptoms for changes.
Such multi-modal analyses may lead to quick response to the natural disaster.


\appendix
\section{Appendix}
\subsection{Additional use case of GRASP EARTH Building 1}
\begin{figure}[h]
    \centering
    \includegraphics[width=0.98\linewidth]{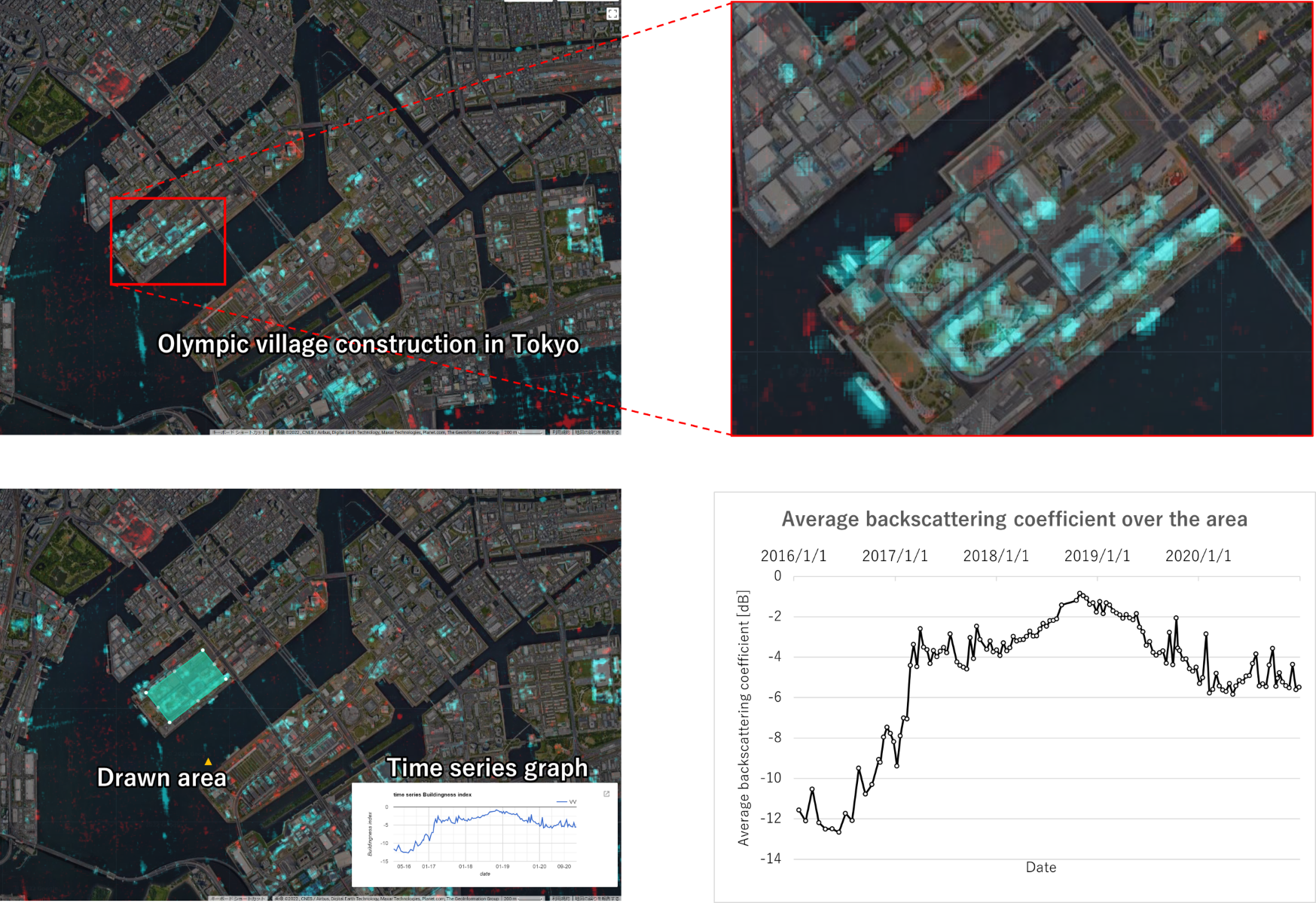}
    \caption{Top left picture shows GRASP EARTH Building screenshot of the Olympic Village in Tokyo, which compares SAR images of January 2016 and January 2021. Top right picture shows the excerpt of red box in top left picture. Bottom left picture draws arbitrary AoI and get time series average of a backscatter coefficient of the area. Bottom right picture shows the graph excepted from bottom left picture.}
    \label{fig:ge_building_tokyo_bay}
\end{figure}

GRASP EARTH Building is used for urban development monitoring. 
Figure~\ref{fig:ge_building_tokyo_bay} shows the screenshot of GRASP EARTH Building in Tokyo bay area, where the Olympic village were built from 2017 to 2020. 
SAR images which were taken in January 2016 and January 2021 are processed and the processed layer is put on the Google map aerial view. 
There are several blue dots over the Olympic village, which means the amount of artifacts has been increased during the period.
Time series signal transition is also available. 
Users can draw arbitrary shapes by selecting button from control panel. 
After drawing the polygon, a graph which shows the average backscatter coefficient over the drawn area emerges as shown in bottom left picture of Figure~\ref{fig:ge_building_tokyo_bay}. 
Bottom right picture of Figure~\ref{fig:ge_building_tokyo_bay} shows the enlarged graph.
The average backscatter coefficient increases from September 2016 to March 2017. 
This signal suggests that the construction started during this period.

\subsection{Additional use case of GRASP EARTH Building 2}
\begin{figure}[h]
    \centering
    \includegraphics[width=0.98\linewidth]{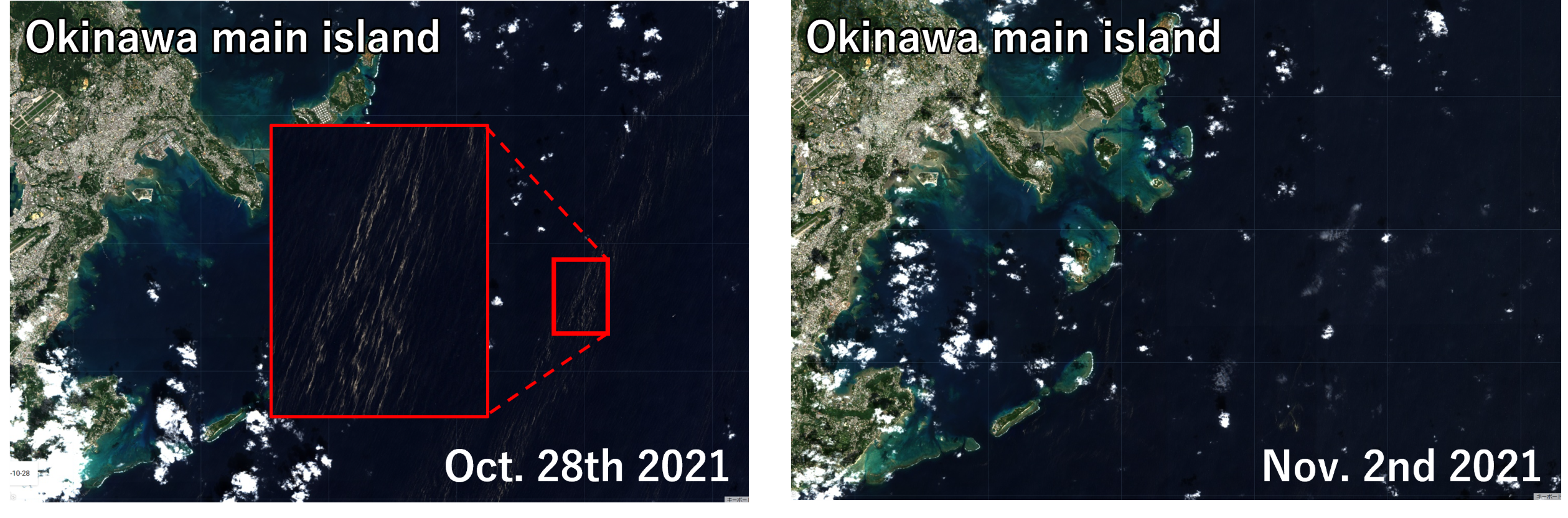}
    \caption{Screenshot of GRASP EARTH Pumice viewer. In this application, window is divided into two and users can compare different two RGB images. The red box in left picture show the area where pumice rafts are detected.}
    \label{fig:GE_pumice_viewer}
\end{figure}

GRASP EARTH Pumice viewer is shown in Figure~\ref{fig:GE_pumice_viewer}.
Cloudless RGB image creation method, which is implemented in GRASP EARTH Color, is not implemented in GRASP EARTH Pumice viewer so that users can detect the exact place of the pumice raft on the input date. 
The pumice raft can be seen in the offshore of Okinawa main island on October 28th, 2021 while the rafts are not seen on November 2nd, 2021. 

\begin{figure}[h]
    \centering
    \includegraphics[width=0.98\linewidth]{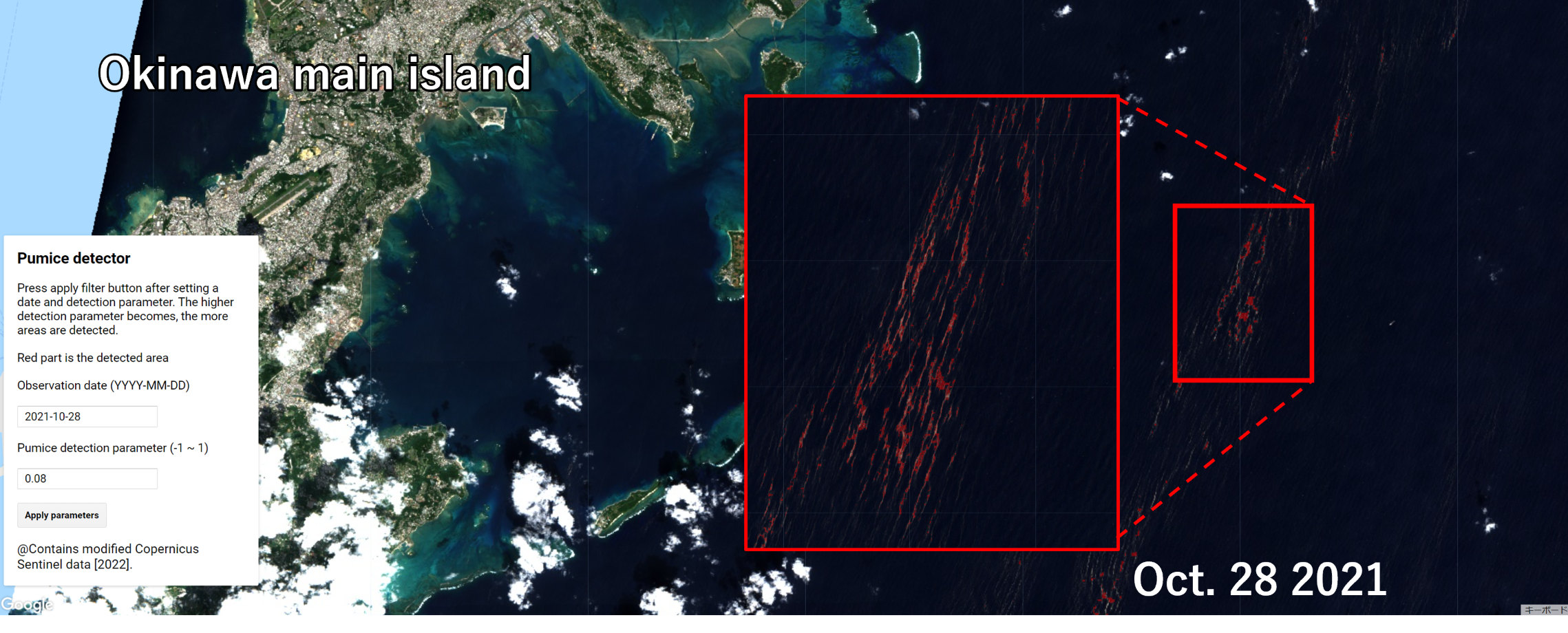}
    \caption{Screenshot of GRASP EARTH pumice detector. Estimated pumice raft area is colored in red.}
    \label{fig:GE_pumice_detector}
\end{figure}

GRASP EARTH Pumice detector, which is shown in Figure~\ref{fig:GE_pumice_detector} was also used for checking places of pumice rafts on daily basis. 
Pumice rafts might emerge everywhere in the Pacific ocean but its width was only dozens of meters at that time.
Therefore, finding pumice rafts required high resolution satellite image. 
GRASP EARTH Pumice detector was created so that it can take over the arduous task.

Deep Learning based segmentation methods and other pixel based classification methods are compared but since the shape and the pixel value changed drastically during the period, we adapted the simple method as described in Section~\ref{sec:overview}. 
This application have been used for checking the rough location of pumice rafts.

Seamless integration of fast satellite images fetching, calculating tools and visualizer enabled these applications to be widely used from media to local people.
GRASP EARTH Pumice viewer and detector could never been created without GEE.

\end{document}